\documentclass[10pt,twocolumn,letterpaper]{article}

\usepackage{iccv}
\usepackage{times}
\usepackage{epsfig}
\usepackage{graphicx}
\usepackage{amsmath}
\usepackage{amssymb}
\usepackage{enumitem}

\usepackage{tikz,pgfplots}
\usepackage{subfig}
\definecolor{note}{rgb}{0.1,0.1,1}
\definecolor{JC}{rgb}{0.6,0.2,0.2}

\usepackage[pagebackref=true,breaklinks=true,letterpaper=true,colorlinks,bookmarks=false]{hyperref}

\iccvfinalcopy

\def\iccvPaperID{1310}

\ificcvfinal\fi
\begin{document}

\title{Video Object Segmentation using Supervoxel-Based Gerrymandering}

\author{Brent A. Griffin and Jason J. Corso\\
University of Michigan\\
Ann Arbor\\
{\tt\small \{griffb,jjcorso\}@umich.edu}
}

\maketitle

\begin{abstract}
    Pixels operate locally. Superpixels have some potential to collect information across many pixels; supervoxels have more potential by implicitly operating across time.   In this paper, we explore this well established notion thoroughly analyzing how supervoxels can be used in place of and in conjunction with other means of aggregating information across space-time.  Focusing on the problem of strictly unsupervised video object segmentation, we devise a method called supervoxel gerrymandering that links masks of foregroundness and backgroundness via local and non-local consensus measures.  We pose and answer a series of critical questions about the ability of supervoxels to adequately sway local voting; the questions regard type and scale of supervoxels as well as local versus non-local consensus, and the questions are posed in a general way so as to impact the broader knowledge of the use of supervoxels in video understanding.  
    We work with the DAVIS dataset and find that our analysis yields an unsupervised method that outperforms all other known unsupervised methods and even many supervised ones.

\end{abstract}

\section{Introduction}
\label{sec:intro}

Video understanding remains a focus area in vision.
A critical sub-problem, video object segmentation, supports learning object class models~\cite{OnReVeECCV2014,TaSuYaCVPR2013}, scene parsing~\cite{LiHeCVPR2015,TiLaIJCV2012} and action recognition~\cite{LuXuCoCVPR2015, SoIdShICCV2015, SoIdShCVPR2016}, and many video editing applications~\cite{ChChChACMM2012}. 
Despite its utility, however, due to the diversity of characteristics videos exhibit and the attendant challenges they present to the assumptions of many segmentation frameworks, finding general solutions for video object segmentation remains a largely unsolved problem.

In an attempt to address some of the challenges in video object segmentation, Faktor and Irani \cite{NLC} relate superpixels in a high-dimensional region space across many frames, thereby mitigating issues arising from foreground objects with inconsistent optical flow or visual saliency properties.
However, while superpixels are restricted to individual frames and {\textit{must} be related across videos by a manufactured coordinate space, supervoxels exhibit an inherent benefit of existing across many frames of video.
To this end, we introduce a video segmentation framework that, starting from an initial pixel-level estimate of ``foregroundness," builds an internal consensus within the bounds of each supervoxel, and then relates this local consensus across the entire video for a final determination of the foreground segmentation.
We conceptualize this approach to swaying the ``foregroundness" of a certain pixel akin to gerrymandering in voting.

Our method has two parts.
First, to build an initial quantitative estimate of ``foregroundness," we use a combination of motion and visual saliency cues that are weighted using a straightforward statistical measure of ``outlierness" \cite{Tukey}.
Second, using the initial estimate we build an internal or local consensus in supervoxels that is then relayed across the video amongst a set of nearest neighbors, which forms an additional non-local consensus. 
Among the many benefits of extending consensus-based segmentation to supervoxels, we find that a three-dimensional feature space forms a reliable non-local consensus, which is a dramatic simplification of the 176-dimensional feature space used to relate superpixels in~\cite{NLC}.

\begin{figure}
\centering
\includegraphics[width=0.475\textwidth]{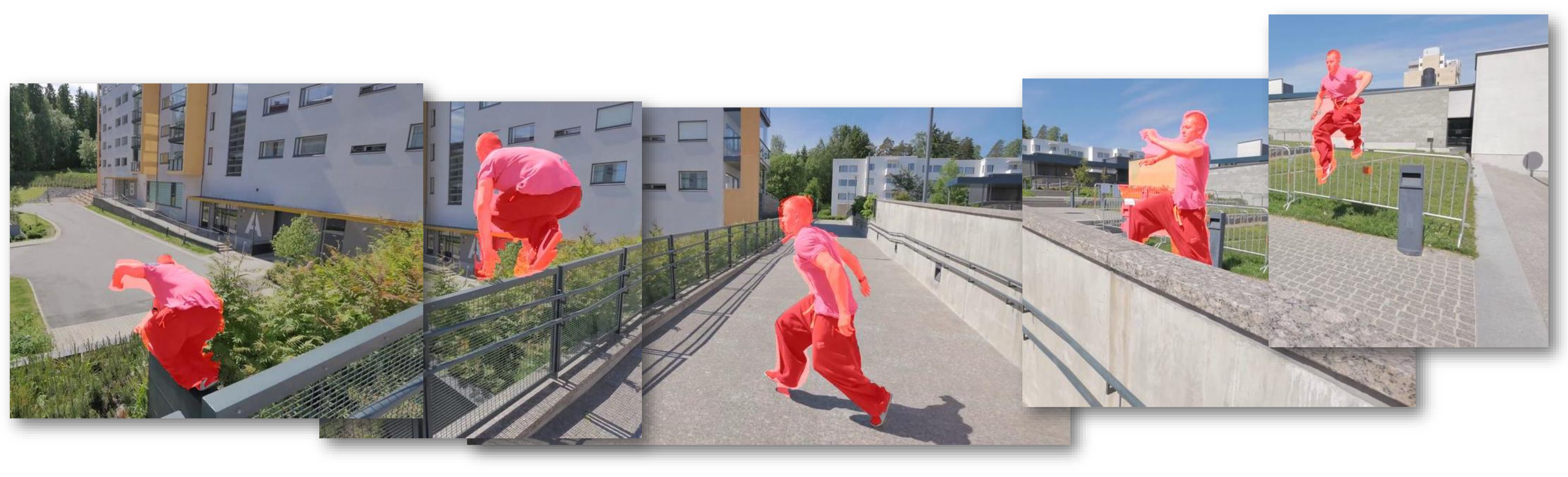}
\caption{Video Object Segmentation of Parkour video in DAVIS dataset. Characteristics of this video include a moving camera and a dynamic foreground object.}
\label{fig:parkour}
\end{figure}

However, we find that there are many unanswered question from the outset of our work.
To address this, we design a system of experiments to explicitly answer the following:
\begin{itemize}[itemsep=0pt]
\item Given the temporal consistency of supervoxels and their inherent reach through multiple frames compared to superpixels, how useful is the internal or local consensus of supervoxels relative to non-local consensus for video object segmentation?
\item Given the availability of multiple methods for generating supervoxels, which methods are the most useful?
\item Given the range of hierarchy levels available with supervoxel methods, which levels are the most effective?
\end{itemize}
\noindent In fact, we believe these questions have broader purview in video understanding than strictly the problem in focus here.

We initiate our consideration of supervoxel selection with a comparative study in \cite{SVXEval}, which found that segmentation by weighted aggregation (SWA) \cite{SWA,SWANature} and hierarchical graph-based (GBH) \cite{GBH} supervoxels perform best and approximately equally-well with respect to spatiotemporal uniformity, object/region boundary detection, region compression, and parsimony.
Xu et al.~\cite{SGBH} later develop a streaming hierarchical graph-based (SG) segmentation method, which only processes each video frame once and can handle arbitrarily long videos.
To answer the supervoxel selection question in the context of this work, we test all three supervoxel methods in our framework and compare their relative performance.

We evaluate the overall efficacy of our supervoxel-based segmentation method 
using multiple annotated video benchmarks, including the Densely Annotated 
VIdeo Segmentation (DAVIS) dataset \cite{DAVIS} (see Figure~\ref{fig:parkour}).
DAVIS has been used in 
propagating annotations 
forward in the video \cite{BVSQ}, training convolutional neural networks 
\cite{FusSeg}, or both \cite{VPNDeep,MaskTrack}.
In this work, we do not consider supervised techniques, whether in the sense of 
requiring prior annotations, training data, or human-in-the-loop functionality.
Although there are indeed many recent developments in supervised methods for 
many vision problems, we remain interested in exploring the comparatively more 
transparent and general unsupervised methods; insights to them may later 
influence supervised methods as well.
Despite the simplicity of this approach, we are able to generate more accurate 
video object segmentations on DAVIS than many supervised techniques 
\cite{SEA,TSP,JMP,GBH,BVSQ,FCP}, and, to the best of the authors' knowledge, 
all current results from unsupervised techniques 
\cite{MSG,NLC,TRC,KEY,FST,CVOS,SAL}.
In our analysis, we thoroughly experiment with all permutations of method configurations to allow us to systematically answer the three questions we posed earlier.  

\paragraph{Main contributions}
Our paper is primarily an analysis paper that asks and answers important 
questions about the use of supervoxels as a means for consensus in video object 
segmentation.  This analysis, which carefully elucidates when certain 
supervoxel methods are more appropriate than others, is the main contribution 
of the paper.  However, a secondary contribution is the actual method, 
supervoxel gerrymandering, which we developed to support this analysis.  
Through the analysis our method is able to achieve the highest performance 
among all known unsupervised methods and higher performance than many 
supervised methods on the DAVIS video object segmentation benchmark 
for single objects, which is our focus.

\paragraph{Paper organization}
The remainder fo this paper is organized as follows.
Section~\ref{sec:approach} derives the supervoxel gerrymandering method and saliency outliers for scaling video data.
Section~\ref{sec:imp} defines our motion and visual saliency-based and supervoxel consensus-based implementations of video object segmentation.
Section~\ref{sec:results} presents our experimental results on the DAVIS and SegTrackv2 datasets with discussion on our findings, and Section~\ref{sec:conclude} provides concluding remarks.

\begin{figure*}
\centering
\includegraphics[width=0.975\textwidth]{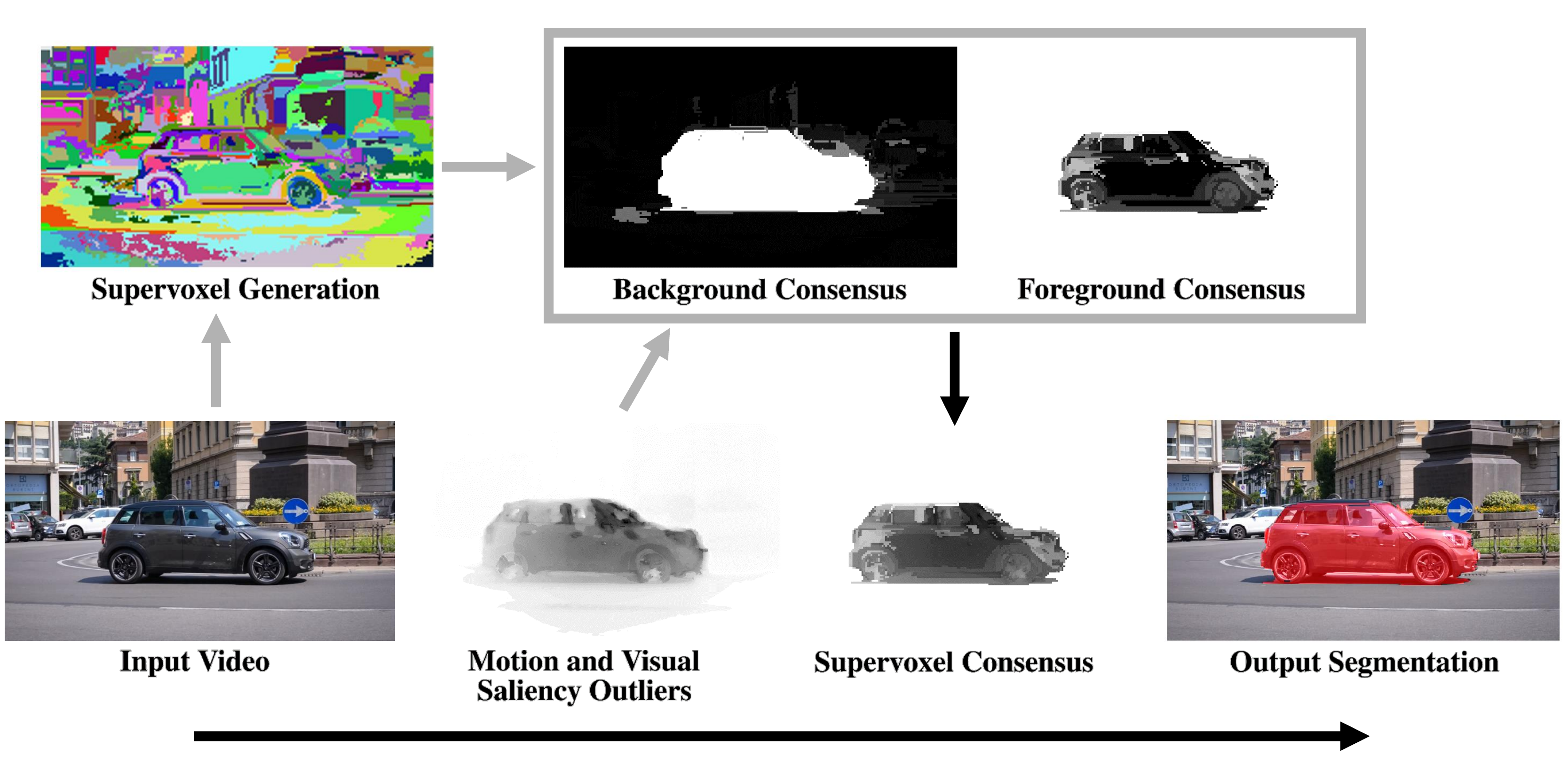}
\caption{Our segmentation framework uses motion and visual saliency measures as an initial estimate of ``foregroundness" ($f_{0p}$ in \eqref{eq:f0p}). 
Using supervoxels generated from the input video, we refine this initial estimate with a supervoxel-based consensus for both background and foreground elements ($f_p$ in \eqref{eq:fp}), which determines our output segmentation ($m_p$ in \eqref{eq:mask}).}
\label{fig:SVXCon}
\end{figure*}

\section{Approach}
\label{sec:approach}
This section derives the foundational concepts of our video segmentation method.
While similar components of this methodology appear in other work \cite{NLC,binSVX,Tukey}, this unification of concepts results from a fine grain analysis and allows a precise description of the implementation framework introduced in Section~\ref{sec:imp}.

We use supervoxel gerrymandering in conjunction with a baseline method that generates an initial quantitative estimate of ``foregroundness" and ``backgroundness." The initial estimate result from a weighted combination of motion and visual saliency cues.
The methodology behind the weighted combination is based on the Tukey range test \cite{Tukey} and, as derived in Section~\ref{sec:SO}, is generalizable to many classes of video data.
From the initial estimate, we use the derivation in Section~\ref{sec:SVXCon} to build a local and non-local supervoxel-based consensus, which improves the overall performance of our video object segmentation framework, which is depicted in Figure~\ref{fig:SVXCon}. 

\subsection{Supervoxel Gerrymandering}
\label{sec:SVXCon}

Supervoxel consensus improves an initial video segmentation using 
supervoxel-level local and non-local consensus voting.
Denote an initial mask estimate at each pixel $p$ as $m_{0p} \in \{0,1\}$, 
where $m_{0p}=1$ suggests that $p$ is in a location corresponding to a 
foreground object.
Also we have a set of non-overlapping 
supervoxels, $S \in \mathcal{S}$, that cover the full set of pixels for a given video; the supervoxels are disjoint.  These may be produced by any method and we will make it concrete during the later analysis.

For each supervoxel, the internal or local foreground consensus is defined as
\begin{align}
f_{S} := \frac{1}{N_{S}}\sum_{p\in S} (2m_{0p}-1),
\label{eq:local}
\end{align}
where $N_{S}$ is the number of $p \in S$ and $f_{S} \in [-1,1]$. Positive or negative values in \eqref{eq:local} correspond to a consensus for $S$ belonging to a foreground object or the background respectively.
For each pixel, the supervoxel foreground consensus is defined as
\begin{align}
f^{sc}_{p} := w_0 f_{S_0} + \sum_{i=1}^{N_{\text{NL}}} w_i f_{S_i},
\label{eq:SVXCon}
\end{align}
where $f_{S_0}$ is the local consensus of the supervoxel containing $p$, $N_{\text{NL}}$ is the number of $S_0$-nearest-neighbor supervoxels contributing to the non-local consensus, and scalar weights $w_0,w_1,...,w_{N_{\text{NL}}}$ determine the relative contribution of each supervoxel.

Although an updated foreground mask can be generated directly from \eqref{eq:SVXCon}, we find it more effective in combination with the initial measure used to find $m_{0p}$.
Hence, a final measure is defined as
\begin{align}
f_p := f_{0p} + f^{sc}_{p},
\label{eq:fp}
\end{align}
where the initial measure $f_{0p}$ improves from $f^{sc}_{p}$ adding ``foregroundness" to pixels with a consistent foreground object consensus and subtracting from pixels with a consistent background consensus as shown in Figure~\ref{fig:SVXCon}.
Finally, the updated foreground mask is chosen as
 \begin{align}
 m_{p} := \begin{cases}
  1 & \text{ if } f_p > 0 \\
  0 & \text{otherwise}
  \end{cases}.
  \label{eq:mask}
  \end{align}

\noindent \textbf{Remarks:}
(a) For the current work, weights in \eqref{eq:SVXCon} are chosen such that $\sum_{i=0}^{N_{\text{NL}}}w_i=1$, which, given $f_{S} \in [-1,1]$ in \eqref{eq:local}, implies that $f^{sc}_{p}\in[-1,1]$.
Correspondingly, $f_{0p}$ is scaled in \eqref{eq:fp} such that $f_{0p} \in [0,1]$.
(b) The scale and offset to $m_{0p}$ in \eqref{eq:local} can be modified to change the relative reward/penalty from a foreground/background consensus.

\subsection{Saliency Outliers}
\label{sec:SO}
Based on the assumption that foreground objects in video exhibit distinct behavior relative to the background, the initial estimate of ``foregroundness," $f_{0p}$ in \eqref{eq:fp}, is found using motion and visual saliency data outliers.
Outliers are found in each set of data using a process inspired by Tukey's range test.
Data are divided using a lower and upper outlier thresholds, ${O_1}$ and ${O_3}$ respectively, that are calculated as
\begin{align}
\label{eq:outlierThresh}
{O_1} =& ~{Q_1} - k({Q_3} - {Q_1}) \\
{O_3} =& ~{Q_3} + k({Q_3} - {Q_1}),
\end{align}
where ${Q_1}$ and ${Q_3}$ are the lower and upper quartiles, ${Q_3} - {Q_1}$ is the interquartile range, $k=1.5$ is a constant that scales the outlier thresholds
(for later use in Section~\ref{sec:init}, $Q_2$ will be considered the median).
For a given set of pixel-level data $d_p\in D$, the outliers are defined as
\begin{align}
O := \{ d_p \in D|d_p<{O_1} \vee d_p>{O_3} \},
\label{eq:O}
\end{align}
allowing us to define an ``outlierness" scale for weighing each saliency measure as
\begin{align}
	\alpha := \frac{ \sum_{d_p \in O} |d_p|}{ \sum_{d_p \in D} |d_p| },
	\label{eq:scale}
\end{align}
where $\alpha \in [0,1]$ is proportional to the magnitude of outlier data relative to all data.

\section{Implementation Framework}
\label{sec:imp}

This section describes the fully implemented segmentation framework.
Section~\ref{sec:init} details the initial quantitative measure of ``foregroundness" and attendant mask.
Section~\ref{sec:SVXimp} describes the supervoxel consensus implementation that improves the initial foreground estimate for the final output mask.
For the sake of completeness and repeatability, source code is provided at \cite{codeSVXGRM}.

\subsection{Initial Segmentation via Saliency Outliers}
\label{sec:init}

This initial estimate of ``foregroundness," $f_{0p}$, is found using a weighted combination of motion and visual saliency measures.
For motion saliency, $x$ and $y$ optical flow components are found using \cite{optFlow}.
Using the $x$ and $y$ flow components, the flow magnitude (i.e., $|x^2 + y^2|$) and flow angle (i.e., $\arctan(\frac{y}{x})$) are also calculated.
For each flow measure, the outlier thresholds and scales are calculated on a frame-to-frame basis using \eqref{eq:outlierThresh}-\eqref{eq:scale}.
The pixel-level measure of motion saliency, $d^{ms}_{p}$, is defined for each flow component as
\begin{align}
d^{ms}_{p} := \begin{cases}
0 & \text{if} ~d_p \notin O \vee \alpha_f < 0.5 \\
\alpha_f|d_{fp} - Q_{f2}| & \text{otherwise}
\end{cases},
\label{eq:motSal}
\end{align}
where $d_{fp}$ is a flow component with corresponding frame-to-frame median $Q_{f2}$, outlier scale $\alpha_f$, and a 0.5 minimum scale requirement.
The intuition behind \eqref{eq:motSal} is simple:
whether a foreground object is moving with a fixed camera or vice versa, the foreground object's deviation from the frame's median optical flow will generally be salient; the absolute value enables a positive ``foregroundness" contribution regardless of a flow component's sign; and the minimum scale requirement will remove the influence of less reliable flow components.

For video segmentation, we have found that in most cases visual saliency is less reliable than optical flow.
However, the product of visual saliency and optical flow is beneficial for videos with dynamic background elements.
Thus, the pixel-level measure of visual saliency, $d^{vs}_p$, is defined as
\begin{align}
d^{vs}_p := {d_{vp}}^{k} \sum_{i=1}^{4} \text{max}(\alpha_{f_i},0.5) |d_{f_ip} - Q_{f_i2}|,
\label{eq:visSal}
\end{align}
where $d_{vp} \in [0,1]$ is a pixel-level visual saliency-based scale (found using \cite{visSal}), $k$ is an exponential scale that adjusts the relative sharpness of ${d_{vp}}^{k} \in [0,1]$, $d_{f_ip}$ is the $i$th flow component with corresponding median $Q_{f_i2}$ and outlier scale $\alpha_{f_i}$, and the minimum applied scale of 0.5 ensures that visual saliency measures are available even if $\alpha_{f_i}=0~ \forall i$.
Three visual saliency measures are used altogether, with $k=\{1,\frac{1}{2},\frac{1}{3}\}$.

Finally, the initial foreground measure, $f_{0p}$, is defined as
\begin{align}
f_{0p} := \sum_{i=1}^{N_D} d^i_{p},
\label{eq:f0p}
\end{align}
where $d^i_{p}$ is the $i$th component of $N_D=7$ saliency measures (four from \eqref{eq:motSal} and three from \eqref{eq:visSal}).
Using \eqref{eq:f0p}, the initial foreground mask, $m_{0p}$, is found using \eqref{eq:mask}, but with the following threshold: $f_{0p} > \beta_p \delta_p$,
where $\beta_p\in \mathbb{R}$ is the sum of the mean and standard deviation of $f_{0p}$ in the current frame and $\delta_p$ is a previous-mask threshold scale defined as
\begin{align}
\delta_p := \begin{cases}
\frac{1}{2} & \text{if} ~ m_{0p,k-1}=1 \\
1 & \text{otherwise}
\end{cases},
\label{eq:delta}
\end{align}
where $m_{0p,k-1}$ represents $m_{0p}$ from the previous frame. In simple words, if $f_{0p}$ is greater than the pixel-level mean and standard deviation of ``foregroundness" over the current frame, $p$ is considered a foreground object location. 
In addition, from \eqref{eq:delta}, wherever $p$ corresponds to a mask position in the previous frame, a half-threshold discount is applied, which generally encourages frame-to-frame segmentation continuity and gradually increasing accuracy of the initial mask.
Finally, the initial mask assumes a single foreground object hypothesis.
Accordingly, the mask is restricted within each frame to contain a single continuous segment exhibiting the greatest combined values of $f_{0p}$.

\noindent \textbf{Remark:}
In the remainder of the paper, the initial foreground mask will be referred to as MVSO (motion and visual saliency outliers).

\subsection{Final Segmentation via Supervoxel Consensus}
\label{sec:SVXimp}
Given the initial foreground estimation from Section~\ref{sec:init}, we now describe our implementation of supervoxel consensus for segmentation.
First, we determine the relative consensus weight of each supervoxel used in \eqref{eq:SVXCon}.
The local consensus weight, $w_0$, changes based on the overall consensus configuration implemented.
If implementing only local consensus, $w_0=1$ and all of the non-local weights are scaled to zero.
If implementing only non-local consensus, $w_0=0$ and all non-local weights are uniformly scaled $s.t. \sum_{i=1}^{N_{\text{NL}}}w_i=1$.
Finally, if implementing both local and non-local consensus, $w_0=\frac{1}{3}$ and non-local weights are uniformly scaled $s.t. \sum_{i=1}^{N_{\text{NL}}}w_i=\frac{2}{3}$.

In \cite{NLC}, non-local consensus weights are based on nearest neighbor regions for superpixels that use a 176-dimensional space, which consists of RGB and LAB histograms, HOG descriptors, and relative spatial coordinates.
In the current work, non-local weights are based on supervoxel regions that only use a 3 dimensional space, which consists of the mean LAB color of each supervoxel.
Specifically, the nominal non-local consensus weights $w_1,...,w_{N_{\text{NL}}}$ in \eqref{eq:SVXCon} are chosen as 
\begin{align}
w_i = \frac{1}{R(S_0,S_i)^2},
\label{eq:NNW}
\end{align}
where $R\in \mathbb{R}$ calculates the city-block distance between the mean-LAB region space of local supervoxel $S_0$ and the $i$th nearest neighbor $S_i$.
Because $R$ is squared, the influence of supervoxels outside of the primary ``clique" drops off quickly.
Furthermore, the LAB region space is normalized for the consensus weight calculation in \eqref{eq:NNW} such that the minimum and maximum video-wide LAB-pixel values correspond to 0 and 1 respectively. This normalization ensures that all three region space dimensions are meaningful, even if their original differences are minor.

Selecting the number of nearest neighbors used for non-local consensus is chosen within the context of the changing number of supervoxels between each video and supervoxel hierarchy. Specifically, $N_{\text{NL}}$ in \eqref{eq:SVXCon} is chosen as
$ N_{\text{NL}} = \frac{1}{N_\mathcal{S}}$,
where $N_\mathcal{S}$ is the total number of supervoxels in the video, $S\in \mathcal{S}$.
Scaling the number of nearest neighbors in this manner keeps a relatively consistent ratio of consensus to non-consensus neighbors for varying sizes of videos and supervoxels.

Given the full definition of the implementation consensus terms, $f^{sc}_p$ is calculated using \eqref{eq:SVXCon} and is combined with $f_{0p}$ from Section~\ref{sec:init} for the final measure of ``foregroundness," $f_p$ in \eqref{eq:fp}. 
Using \eqref{eq:mask}, values of $f_p>0$ determine the locations of the foreground object proposal, $m_p$. 
As a final processing step, $m_p$ is restricted within each frame to contain at most two continuous segments exhibiting the greatest combined values of $f_p$.
In this sense, the final foreground segmentation of our framework roughly approximates a single object hypothesis that accounts for limited instances of occlusions.

\section{Results}
\label{sec:results}

\setlength{\tabcolsep}{4.75pt}
\begin{table*}
\centering
\begin{tabular}{ r | c | c | c | c | c | c | c | c | c | c}
\hline
& \multicolumn{10}{c}{Configuration ID} \\
\multicolumn{1}{c|}{Configuration} & \multicolumn{1}{c}{MVSO} & \multicolumn{1}{c}{SWA$^{06}$} & \multicolumn{1}{c}{SWA$^{08}_{\text{L}}$} & \multicolumn{1}{c}{SWA$^{05}_{\text{NL}}$} &
\multicolumn{1}{c}{GBH$^{00}$} & \multicolumn{1}{c}{GBH$^{02}_{\text{L}}$} & \multicolumn{1}{c}{GBH$^{00}_{\text{NL}}$} &
\multicolumn{1}{c}{SG$^{01}$} & \multicolumn{1}{c}{SG$^{00}_{\text{L}}$} & \multicolumn{1}{c}{SG$^{00}_{\text{NL}}$} \\
\hline
Supervoxel Method & None & SWA & SWA & SWA & GBH & GBH & GBH & SG & SG & SG \\
Hierarchy Level & N/A & 6 & 8 & 5 & 0 & 2 & 0 & 1 & 0 & 0 \\
Local Consensus & N/A & Yes & Yes & No & Yes & Yes & No & Yes & Yes & No \\
Non-Local Consensus & N/A & Yes & No & Yes & Yes & No & Yes & Yes & No & Yes \\
\hline
\multicolumn{1}{c|}{Measure} & \multicolumn{10}{c}{DAVIS Results} \\
\hline
\multicolumn{1}{r|}{Rank} & \multicolumn{1}{c}{6th } & \multicolumn{1}{c}{\bf 1st }	& \multicolumn{1}{c}{ 4th}	& \multicolumn{1}{c}{5th }	& \multicolumn{1}{c}{\bf 3rd }	& \multicolumn{1}{c}{\bf 2nd }	& \multicolumn{1}{c}{ 7th}	& \multicolumn{1}{c}{9th }	& \multicolumn{1}{c}{8th }	& \multicolumn{1}{c}{10th }  \\
\hline
$\mathcal{J}$ Mean $\uparrow$ & 0.586 & 0.676 & 0.647	& 0.616	& 0.648	& 0.653	& 0.544	& 0.495	& 0.537	& 0.282\\
$\mathcal{F}$ Mean $\uparrow$ & 0.475 & 0.639 & 0.615	& 0.592	& 0.61	& 0.612	& 0.526	& 0.488	& 0.51	& 0.326 \\
$\mathcal{T}$ Mean $\downarrow$ & 0.307 & 0.310 & 0.31	& 0.431	& 0.33	& 0.318	& 0.522	& 0.425	& 0.364	& 0.78\\
\hline
Supervoxel Volume & N/A & 3617 & 24651 &	1342 &	9745 &	32988 &	9745 &	292317 & 155447 & 155447\\
\hline

\end{tabular}
\caption{Results for each combination of supervoxel and consensus method on the DAVIS dataset. Result measures consist of region similarity ($\mathcal{J}$), contour accuracy ($\mathcal{F}$), and temporal (in-)stability ($\mathcal{T}$). For rows with an upward pointing arrow higher numbers are better (e.g., $\mathcal{J}$ mean), and vice versa for rows with downward pointing arrows. Rank is based on $\mathcal{J}$. Supervoxel volume is calculated as an average over the entire DAVIS dataset.}
\label{tab:Config}
\end{table*}

\setlength{\tabcolsep}{3.5pt}
\begin{table*}
\centering
\begin{tabular}{ r | c c c c c c | c c c c c c c }
\hline
\multicolumn{14}{c}{DAVIS Results for State-of-the-Art Unsupervised Methods}\\
\hline
\multicolumn{1}{c |}{} & \multicolumn{6}{c|}{Current Results} & \cite{NLC} & \cite{FST} & \cite{KEY} & \cite{MSG} & \cite{CVOS} & \cite{TRC} & \cite{SAL}  \\
\multicolumn{1}{c |}{Measure} & SWA$^{06}$ & GBH$^{02}_{\text{L}}$ & GBH$^{00}$ & SWA$^{08}_{\text{L}}$ & SWA$^{05}_{\text{NL}}$ & MVSO & NLC & FST & KEY & MSG & CVOS & TRC & SAL  \\
\hline
$\mathcal{J}$ Rank & \bf 1st & \bf 2nd & \bf 3rd & 4th & 6th & 7th & 5th & 8th & 9th & 10th & 11th & 12th & 13th \\
$\mathcal{F}$ Rank & \bf 1st & \bf 3rd & 4th & \bf 2nd & 6th & 12th & 5th & 7th & 9th & 8th & 10th & 11th & 13th \\
$\mathcal{T}$ Rank & 6-7th & 8th & 10th & 6-7th & 12th & 5th & 11th & 4th & \bf 1st & \bf 3rd & \bf 2nd & 9th & 13th \\
\hline
	                         Mean $\uparrow$     &\bf\ 0.676 &   \ 0.653 &   \ 0.648 &   \ 0.647 &   \ 0.616 &   \ 0.586 &   \ 0.641 &   \ 0.575 &   \ 0.569 &   \ 0.543 &   \ 0.514 &   \ 0.501 &   \ 0.426 \\
$\mathcal{J}$ Recall $\uparrow$   &\bf\ 0.847 &   \ 0.822 &   \ 0.822 &   \ 0.823 &   \ 0.756 &   \ 0.759 &   \ 0.731 &   \ 0.652 &   \ 0.671 &   \ 0.636 &   \ 0.581 &   \ 0.560 &   \ 0.386 \\
	                         Decay $\downarrow$  &   \ 0.040 &   \ 0.024 &   \ 0.036 &   \ 0.033 &   \ 0.032 &\bf\ 0.023 &   \ 0.086 &   \ 0.044 &   \ 0.075 &   \ 0.028 &   \ 0.127 &   \ 0.050 &   \ 0.084 \\
\hline
	                         Mean $\uparrow$     &\bf\ 0.639 &   \ 0.612 &   \ 0.610 &   \ 0.615 &   \ 0.592 &   \ 0.475 &   \ 0.593 &   \ 0.536 &   \ 0.503 &   \ 0.525 &   \ 0.490 &   \ 0.478 &   \ 0.383 \\
$\mathcal{F}$ Recall $\uparrow$   &\bf\ 0.785 &   \ 0.755 &   \ 0.768 &   \ 0.755 &   \ 0.708 &   \ 0.488 &   \ 0.658 &   \ 0.579 &   \ 0.534 &   \ 0.613 &   \ 0.578 &   \ 0.519 &   \ 0.264 \\
	                         Decay $\downarrow$  &   \ 0.057 &   \ 0.040 &   \ 0.060 &   \ 0.046 &   \ 0.054 &\bf\ 0.014 &   \ 0.086 &   \ 0.065 &   \ 0.079 &   \ 0.057 &   \ 0.138 &   \ 0.066 &   \ 0.072 \\
	                         \hline
$\mathcal{T}$ Mean $\downarrow$   &   \ 0.310 &   \ 0.318 &   \ 0.330 &   \ 0.310 &   \ 0.431 &   \ 0.307 &   \ 0.356 &   \ 0.276 &\bf\ 0.190 &   \ 0.250 &   \ 0.243 &   \ 0.327 &   \ 0.600 \\
\hline

\end{tabular}
\caption{Comparison of our segmentation method with other state-of-the-art unsupervised methods on the DAVIS dataset. 
Bold text indicates the best performance for a specific measure.
Multiple combinations of supervoxel and consensus method achieve state-of-the-art results.
The current work is only outperformed in temporal (in-)stability ($\mathcal{T}$).
Object recall measures the fraction of sequences scoring higher than 0.5, and decay quantifies the performance loss (or gain) over time \cite{DAVIS}.
MVSO exhibits the best decay performance for region similarity and contour accuracy, which is likely the result of the previous-mask threshold discount \eqref{eq:delta} implemented on MVSO, which encourages gradually increasing accuracy of the initial mask.}
\label{tab:DAVISCompare}
\end{table*}

We evaluate our approach on two datasets: the Densely Annotated VIdeo Segmentation (DAVIS) \cite{DAVIS} and the Georgia Tech Segmentation and Tracking Dataset (SegTrackv2) \cite{SegTrackv2,SegTrack}.
The DAVIS dataset includes videos from 50 diverse scenarios that collectively test the assumptions of many segmentation frameworks.
Furthermore, all of the DAVIS dataset's ground truth annotations match the single object hypothesis.
The SegTrackv2 dataset has fewer videos than DAVIS, and only a subset match the single object hypothesis.
However, SegTrackv2 provides an additional challenge by using videos with different resolutions, which span from 76,800 to 230,400 pixels per frame.

Three different, accepted, measures are used to evaluate the performance of our video foreground object segmentation:
region similarity $\mathcal{J}$, contour accuracy $\mathcal{F}$, and temporal (in-)stability $\mathcal{T}$, which are all calculated using the definitions provided in \cite{DAVIS}.
Region similarity (also known as the intersect over union or Jaccard index \cite{jaccard}) provides a straightforward and scale-invariant evaluation of the number of mislabeled foreground pixels with respect to a ground truth annotation.
Given a foreground mask $M$ and ground truth annotation $G$, $\mathcal{J}=\frac{M\cap G}{M \cup G}$.
Contour accuracy evaluates the boundary of a segmentation by measuring differences between the closed set of contours for $M$ and $G$. 
Finally, temporal stability is a measure based on the consistency of a mask between frames.
While temporal stability is important for some applications, such as video editing, it is shown to be generally uncorrelated with region similarity and contour accuracy in \cite{DAVIS}.

To test the efficacy of our segmentation approach and establish best practices for implementation, we evaluate the relative changes in performance for all permutations of the following configuration options:
\begin{itemize}
\item Supervoxel methods: segmentation by weighted aggregation (SWA), hierarchical graph-based segmentation (GBH), and streaming hierarchical graph-based segmentation (SG).
\item Supervoxel hierarchy levels: 5-12 for SWA and 0-20 for GBH and SG.
\item Consensus methods: local consensus, non-local consensus, and both.
\end{itemize}
Following this evaluation in Section~\ref{sec:eval}, Section~\ref{sec:discuss} provides insights into the questions we posed at the beginning of our study.

\noindent \textbf{Remarks:}
(a) For simplicity, the SWA, GBH, and SG supervoxels are generated using the standard settings provided in the supervoxel library LIBSVX \cite{LIBSVX}.
(b) To improve computation time, supervoxels are processed on a scaled-down resolution for both data sets and scaled back for segmentation.
The reduction scale is 1:4 for the DAVIS dataset and 1:2 for SegTrackv2, which uses lower resolution images than DAVIS.

\subsection{Dataset Evaluation}
\label{sec:eval}

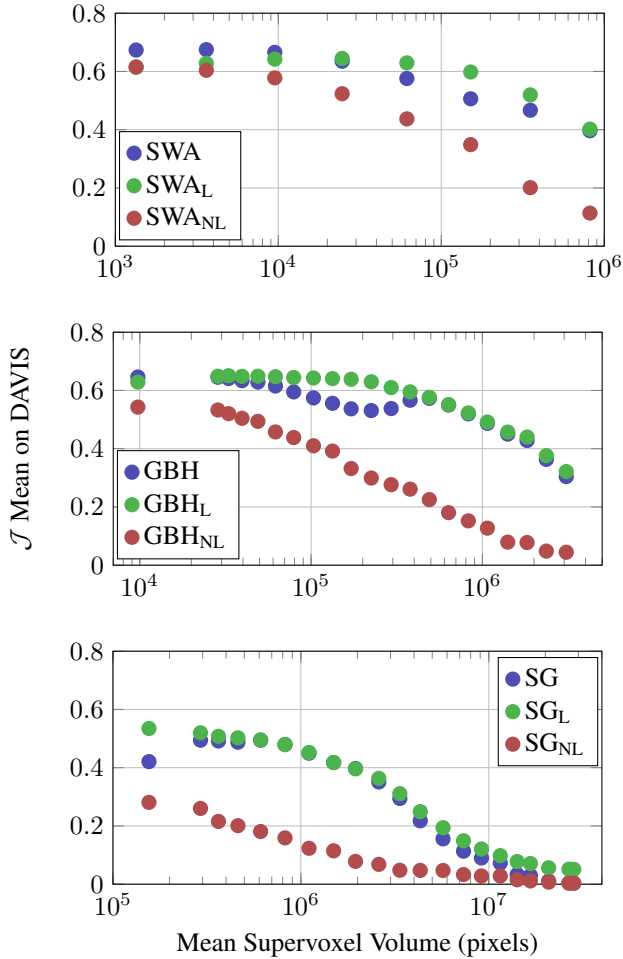
\begin{figure}
  \centering
  \subfloat{
%
%
\begin{tikzpicture}

\begin{axis}[%
width=6.5cm,
height=3.1cm,
scale only axis,
xmode=log,,
xmin=1000,
xmax=1000000,
xminorticks=true,
xmajorgrids,
ymin=0,
ymax=0.8,
ylabel={$~ $},
y label style={at={(-0.045,0.5)}},
ymajorgrids,
legend style={at={(0.0075,0.025)},anchor=south west},
legend cell align=left,
]
\addplot [color=blue!40!gray,only marks,line width=0.75pt,mark size=2.5pt,mark=*,mark options={solid}]
  table[row sep=crcr]{%
1341.79842	0.6735107\\
3617.205993	0.67497364\\
9517.520009	0.665815422\\
24650.56567	0.635080996\\
61510.25407	0.575846534\\
151473.1123	0.506201476\\
351869.2174	0.466816186\\
816766.782	0.396385969\\
};
\addlegendentry{SWA};

\addplot [color=green!40!gray,only marks,line width=0.75pt,mark size=2.5pt,mark=*,mark options={solid}]
  table[row sep=crcr]{%
1341.79842	0.61431004\\
3617.205993	0.62807332\\
9517.520009	0.64225126\\
24650.56567	0.6446147\\
61510.25407	0.62952052\\
151473.1123	0.598203712\\
351869.2174	0.519537072\\
816766.782	0.4020092\\
};
\addlegendentry{SWA$_{\text{L}}$};

\addplot [color=red!40!gray,only marks,line width=0.75pt,mark size=2.5pt,mark=*,mark options={solid}]
  table[row sep=crcr]{%
1341.79842	0.615519192\\
3617.205993	0.604043716\\
9517.520009	0.578152674\\
24650.56567	0.523616462\\
61510.25407	0.437197322\\
151473.1123	0.348881615\\
351869.2174	0.201076057\\
816766.782	0.113691625\\
};
\addlegendentry{SWA$_{\text{NL}}$};

\end{axis}
\end{tikzpicture}%
  \label{fig:notName1}}
  
  \subfloat{
%
%
\begin{tikzpicture}

\begin{axis}[%
width=6.5cm,
height=3.1cm,
scale only axis,
xmode=log,
xmin=7000,
xmax=5000000,
xminorticks=true,
xmajorgrids,
ymin=0,
ymax=0.8,
ylabel={$\mathcal{J}$ Mean on DAVIS},
ymajorgrids,
legend style={at={(0.0075,0.025)},anchor=south west},
legend cell align=left,
]
\addplot [color=blue!40!gray,only marks,line width=0.75pt,mark size=2.5pt,mark=*,mark options={solid}]
  table[row sep=crcr]{%
9744.554837	0.64622028\\
28594.83535	0.64536758\\
32987.99525	0.64094484\\
39513.25503	0.63366168\\
48911.23083	0.62877858\\
61932.55204	0.61530864\\
79147.96021	0.5950106\\
103264.5716	0.574419684\\
133296.3388	0.555912934\\
171316.5114	0.536995524\\
224484.641	0.530932932\\
293345.8161	0.53784098\\
378480.3848	0.56632872\\
489890.5569	0.572842302\\
634531.1828	0.549929348\\
827746.1134	0.51958298\\
1068885.736	0.487060796\\
1405033.333	0.450111604\\
1822745.946	0.428123068\\
2360456	0.363360461\\
3085563.399	0.304437735\\
};
\addlegendentry{GBH};

\addplot [color=green!40!gray,only marks,line width=0.75pt,mark size=2.5pt,mark=*,mark options={solid}]
  table[row sep=crcr]{%
9744.554837	0.62862956\\
28594.83535	0.6488148\\
32987.99525	0.6505791\\
39513.25503	0.64856382\\
48911.23083	0.6489042\\
61932.55204	0.64716416\\
79147.96021	0.64420386\\
103264.5716	0.6427677\\
133296.3388	0.64067998\\
171316.5114	0.63849242\\
224484.641	0.62979658\\
293345.8161	0.60989206\\
378480.3848	0.59507736\\
489890.5569	0.576285506\\
634531.1828	0.55132633\\
827746.1134	0.52286568\\
1068885.736	0.491301652\\
1405033.333	0.45711676\\
1822745.946	0.43980898\\
2360456	0.376088604\\
3085563.399	0.3214642\\
};
\addlegendentry{GBH$_{\text{L}}$};

\addplot [color=red!40!gray,only marks,line width=0.75pt,mark size=2.5pt,mark=*,mark options={solid}]
  table[row sep=crcr]{%
9744.554837	0.542886772\\
28594.83535	0.5329296\\
32987.99525	0.5203166\\
39513.25503	0.50450131\\
48911.23083	0.4939452\\
61932.55204	0.457576286\\
79147.96021	0.43835036\\
103264.5716	0.410082002\\
133296.3388	0.391691109\\
171316.5114	0.331978416\\
224484.641	0.299275185\\
293345.8161	0.276546126\\
378480.3848	0.261151411\\
489890.5569	0.225173209\\
634531.1828	0.18045117\\
827746.1134	0.152223739\\
1068885.736	0.127224044\\
1405033.333	0.079200585\\
1822745.946	0.077737803\\
2360456	0.048340469\\
3085563.399	0.044547701\\
};
\addlegendentry{GBH$_{\text{NL}}$};

\end{axis}
\end{tikzpicture}%
  \label{fig:notName2}}
  
  \subfloat{
%
%
\begin{tikzpicture}

\begin{axis}[%
width=6.5cm,
height=3.1cm,
scale only axis,
xmode=log,
xmin=100000,
xmax=40000000,
xminorticks=true,
xlabel={Mean Supervoxel Volume (pixels)},
xmajorgrids,
ymin=0,
ymax=0.8,
ylabel={$~ $},
y label style={at={(-0.02,0.5)}},
ymajorgrids,
legend style={at={(0.9925,0.975)},anchor=north east},
legend cell align=left,
]
\addplot [color=blue!40!gray,only marks,line width=0.75pt,mark size=2.5pt,mark=*,mark options={solid}]
  table[row sep=crcr]{%
155446.5591	0.420843352\\
292316.5325	0.494380202\\
364267.9012	0.491726866\\
462381.1949	0.487467774\\
610726.0026	0.494209158\\
823414.8837	0.479349182\\
1104737.598	0.449767542\\
1493959.494	0.41759641\\
1953480.828	0.39741381\\
2598667.156	0.350945474\\
3364070.309	0.29425771\\
4317907.317	0.217614221\\
5710780.645	0.155605895\\
7338205.181	0.113301237\\
9137249.032	0.089951757\\
11514419.51	0.072753689\\
14162736	0.033193256\\
16662042.35	0.02972539\\
20827552.94	0.010967536\\
26722143.4	0.004325434\\
28325472	0.003474756\\
};
\addlegendentry{SG};

\addplot [color=green!40!gray,only marks,line width=0.75pt,mark size=2.5pt,mark=*,mark options={solid}]
  table[row sep=crcr]{%
155446.5591	0.534608702\\
292316.5325	0.519792966\\
364267.9012	0.507712032\\
462381.1949	0.502441184\\
610726.0026	0.49555538\\
823414.8837	0.479631812\\
1104737.598	0.452200428\\
1493959.494	0.41791051\\
1953480.828	0.396166034\\
2598667.156	0.363587002\\
3364070.309	0.310720561\\
4317907.317	0.249231813\\
5710780.645	0.194047948\\
7338205.181	0.149039637\\
9137249.032	0.120785378\\
11514419.51	0.098109239\\
14162736	0.077764078\\
16662042.35	0.070900292\\
20827552.94	0.056203369\\
26722143.4	0.051544639\\
28325472	0.050837125\\
};
\addlegendentry{SG$_{\text{L}}$};

\addplot [color=red!40!gray,only marks,line width=0.75pt,mark size=2.5pt,mark=*,mark options={solid}]
  table[row sep=crcr]{%
155446.5591	0.280982983\\
292316.5325	0.260096064\\
364267.9012	0.215514235\\
462381.1949	0.200974671\\
610726.0026	0.180846788\\
823414.8837	0.158965013\\
1104737.598	0.123301035\\
1493959.494	0.114596749\\
1953480.828	0.078311503\\
2598667.156	0.068049996\\
3364070.309	0.047845715\\
4317907.317	0.047443443\\
5710780.645	0.047026918\\
7338205.181	0.033090261\\
9137249.032	0.028193824\\
11514419.51	0.028387851\\
14162736	0.015165564\\
16662042.35	0.011319026\\
20827552.94	0.007178961\\
26722143.4	0.003658057\\
28325472	0.003474756\\
};
\addlegendentry{SG$_{\text{NL}}$};

\end{axis}
\end{tikzpicture}%
  \label{fig:notName3}}
  
  \caption{$\mathcal{J}$ mean on the DAVIS dataset vs. mean supervoxel volume for SWA hierarchy levels 5-12 (top), GBH 0-20 (middle), and SG 0-20 (bottom). Mean supervoxel volumes increase with hierarchy level.}
   \label{fig:hierVsDJ}
\end{figure}

\paragraph{DAVIS}
Here, we present the results of our evaluation experiments using the DAVIS dataset.
Best results for each type of configuration are provided in Table~\ref{tab:Config}.
To help explain the naming convention, SWA$^{06}$ uses SWA hierarchy level 6 and represents the best results for any SWA configuration using both local and non-local consensus, while GBH$^{02}_{\text{L}}$ uses GBH hierarchy level 2 and represents the best results for any GBH configuration using only local consensus.

The best result for region similarity ($\mathcal{J}$) and contour accuracy ($\mathcal{F}$) both come from SWA$^{06}$, which uses both forms of consensus and the second lowest hierarchy available to SWA. GBH$^{02}_{\text{L}}$ has the second best $\mathcal{J}$ and third best $\mathcal{F}$ and uses only local consensus on the third lowest hierarchy available to GBH. GBH$^{00}$ has the third best $\mathcal{J}$, and SWA$^{08}_{\text{L}}$ has the second best $\mathcal{F}$.
The best result using only non-local consensus is SWA$^{05}_{\text{NL}}$, which achieves a middle result of the 5th best $\mathcal{J}$ and uses the lowest SWA hierarchy. The significance of this hierarchy is that it has the smallest supervoxel volume and largest number of supervoxels to form a non-local consensus.
Overall, SG configurations have the worst results and largest supervoxels.
The SG non-local consensus configuration, SG$^{00}_{\text{NL}}$, has the worst performance in all categories.
The initial mask configuration, MVSO from Section~\ref{sec:init}, has a better $\mathcal{J}$ result than all SG configurations and GBH$^{00}_{\text{NL}}$, but the second worst contour.

The region similarity results for all configurations are shown in Figure~\ref{fig:hierVsDJ}.
In all cases, performance eventually decreases with increasing hierarchy level and supervoxel size, although the local consensus configuration groups SWA$_\text{L}$ and GBH$_\text{L}$ hold out against this trend longer than their counterpart configurations.
At the highest hierarchy levels, SWA and GBH configurations, which use both local and non-local consensus, converge on SWA$_\text{L}$ and GBH$_\text{L}$ as the number of non-local consensus neighbors decreases with the number of supervoxels (recall from Section~\ref{sec:SVXimp} that $N_{\text{NL}} = \frac{1}{N_\mathcal{S}}$). 

Finally, the current work is compared with the current state-of-the-art unsupervised segmentation methods in Table~\ref{tab:DAVISCompare} and the current state-of-the-art supervised segmentation methods in Table~\ref{tab:DAVISSuper}, with more discussion provided in Section~\ref{sec:discuss}.

\setlength{\tabcolsep}{7.5pt}
\begin{table*}
\centering
\begin{tabular}{ c | c c c c c c c c c c}
\hline
\multicolumn{11}{c}{Comparison of DAVIS Results with Supervised Methods}\\
\hline
& \cite{VPNDeep} & \cite{MaskTrack} & \cite{FusSeg} & \multicolumn{1}{|c|}{Current Results} & \cite{BVSQ} & \cite{FCP} & \cite{JMP} & \cite{GBH} & \cite{SEA} & \cite{TSP}  \\
Measure & VPND	& MSKT	& FUSG	& \multicolumn{1}{|c|}{SWA$^{06}$} & BVS$_{\text{Q}}$	& FCP	& JMP	& HVS	& SEA	& TSP  \\
\hline
	$\mathcal{J}$ Mean &\ 0.750 &\ 0.748 & \ 0.715 & \multicolumn{1}{|c|}{\ 0.676} &\ 0.67	&\ 0.631	&\ 0.607	&\ 0.596	&\ 0.556	&\ 0.358 \\
	\hline
\end{tabular}
\caption{Comparison of our unsupervised segmentation method with supervised methods on the DAVIS dataset.}
\label{tab:DAVISSuper}
\end{table*}

\setlength{\tabcolsep}{5.5pt}
\begin{table*}
\centering
\begin{tabular}{ r | c c c c c c c | c c c | c c }
\hline
\multicolumn{13}{c}{SegTrackv2 Results}\\
\hline
& \multicolumn{10}{c|}{Unsupervised} & \multicolumn{2}{c}{Supervised} \\
\cline{2-13}
\multicolumn{1}{c|}{Video} & \multicolumn{7}{c|}{Current Results} & \cite{NLC} & \cite{KEY} & \cite{FST} & \cite{FusSeg} & \cite{GBH} \\
\multicolumn{1}{c|}{$\mathcal{J}$ Mean} & GBH$^{\text{ALL}}_{\text{L}}$ &	GBH$^{09}_{\text{L}}$ & SWA$^{\text{ALL}}$ & SG$^{\text{ALL}}_{\text{L}}$ &	SWA$^{05}$ & SG$^{01}_{\text{L}}$ &	MVSO &	NLC &	KEY &	FST &	FUSG &	HVS \\
\hline
\multicolumn{1}{r|}{Rank} & \bf 2nd & \bf 3rd	& 5th	& 6th	& 7th/8th	& 9th	& 12th	& \bf 1st	& 4th	& 11th	& 7th/8th	& 10th  \\
\hline
Birdfall & 0.62	& 0.62	& 0.54	& 0.50	& 0.54	& 0.49	& 0.23	& \bf 0.74	& 0.49	& 0.18	& 0.38	& 0.57 \\
Frog & 0.78	&  0.74	&  0.50	&  0.47	&  0.50	&  0.44	&  0.61	& \bf 0.83	&  0.00	&  0.54	&  0.57	&  0.67 \\
Girl & 0.69	&  0.67	&  0.70	&  0.64	&  0.70	&  0.62	&  0.65	& \bf 0.91	&  0.88	&  0.55 &  	0.67	&  0.32  \\
Monkey & 0.58	&  0.52	&  0.57	&  0.60	&  0.44	&  0.49	&  0.34	&  0.71	&  0.79 &  	0.65	&  \bf 0.80	&  0.62
 \\
Parachute & 0.88	&  0.88	&  0.88	&  0.86	&  0.86	&  0.85	&  0.67	&  0.94	& \bf 0.96	&  0.76	&  0.52	&  0.69
 \\
Soldier &  0.56	&  0.49	&  0.53	&  0.63	&  0.53	&  0.58	&  0.49	& \bf 0.83	&  0.67	&  0.40	&  0.70	&  0.67
\\
Worm &  0.77	&  0.75	&  0.58	&  0.58	&  0.58	&  0.58	&  0.52	&  0.81	& \bf 0.84	&  0.73	&  0.51	&  0.35 \\
\hline
All & 0.70	& 0.67	& 0.62	& 0.61	& 0.59	& 0.58	& 0.50	& \bf 0.82	& 0.66	& 0.54	& 0.59	& 0.56 \\
\hline
\end{tabular}
\caption{Comparison of our segmentation method with other state-of-the-art methods on the SegTrackv2 dataset.
Bold text indicates the best performance for each video.
Methods labeled with an ``ALL" can switch suprevoxel hierarchies for each video, which have different resolutions on the SegTrackv2 dataset.
Selected videos consist of a single foreground object hypothesis.
Results for other methods come from comparative studies in \cite{NLC,FusSeg}.
}
\label{tab:SegTrackCompare}
\end{table*}

\paragraph{SegTrackv2}
Table~\ref{tab:SegTrackCompare} shows the results of an additional evaluation using the SegTrackv2 dataset.
The videos included in our evaluation are those that use a single hypothesis annotation, which is consistent with the DAVIS dataset.
However, SegTrackv2 also presents the challenge of using videos with different video resolutions, spanning from 76,800 to 230,400 pixels per frame.
This is challenging because a configuration with a constant supervoxel hierarchy level will have a dramatically different number of supervoxels from one video to the next.
To remedy this limitation, we introduce an additional set of configurations, labeled with an ``ALL," that can switch supervoxel hierarchy levels for each video, which is shown to provide a slight increase in performance.
Compared with the DAVIS dataset, GBH and SG configuration exhibit an increase in performance, and SWA configurations and MVSO exhibit a decrease in performance.
It is also worth noting that an increase in performance from MVSO would likely improve the results of all other configurations since it forms the initial basis for supervoxel consensus.

\subsection{Discussion}
\label{sec:discuss}

\paragraph{Local vs. Non-Local Consensus}
From Figure~\ref{fig:hierVsDJ}, it is evident that our non-local consensus framework is heavily reliant on using a low-level hierarchy, and quickly loses performance as supervoxel size increases and the attendant number of available neighbors for consensus decreases.
The local consensus framework, on the other hand, exhibits a greater robustness to increasing hierarchy levels, which makes intuitive sense since larger supervoxels will be able to draw on more internal pixels throughout a video for a better informed consensus.
Multiple local consensus configurations achieve the best results of the current work on the SegTrackv2 dataset (see Table~\ref{tab:SegTrackCompare}), and also exhibit some of the best results on the DAVIS dataset.
Still, it is worth noting that the SWA$^{06}$ configuration, which uses both local and non-local consensus, achieves the best results on the DAVIS dataset, which implies that there are merits to using both forms of consensus when using a hierarchy level where both are independently effective (see Figure~\ref{fig:hierVsDJ} top).

All consensus-based methods exhibit a poorer temporal stability ($\mathcal{T}$) relative to the other methods presented in Table~\ref{tab:DAVISCompare}. This likely occurs from discrete consensus-based changes in the segmentation mask that can occur from frame to frame.
However, methods using local supervoxel consensus appear to be less effected by this phenomena than the non-local consensus-based SWA$^{05}_{\text{NL}}$ or non-local superpixel-consensus based NLC.

\paragraph{Choice of Supervoxel Method}
The SWA configuration SWA$^{06}$ exhibits the best performance on the DAVIS dataset (see Table~\ref{tab:Config}).
However, GBH configurations consist of the second and third best performances, and the two best performances of the current work on the SegTrackv2 dataset, which has a variable resolution between videos.
Thus, of the three methods, GBH appears to be the most robust framework with respect to video scale and variability.

SG configurations have the worst performance of the current methods on the DAVIS dataset, however, they exhibit an improvement on the SegTrackv2 dataset and operate in a streaming format, which works well for long videos and is necessary for any eventual real-time application.
In the end, we propose that SG should only be considered if streaming or better runtime performance on longer videos is necessary, while both GBH and SWA methods warrant further investigation for the general video segmentation problem.

\paragraph{Choice of Supervoxel Hierarchy Level}
In Figure~\ref{fig:hierVsDJ}, the overall trend is clear that performance eventually drops when using the highest hierarchy levels.
This intuitive makes sense because larger supervoxels in high hierarchy levels are more likely to have segmented over a meaningful foreground object boundary.
Low hierarchy levels also enable non-local consensus forming with larger groups of relevant supervoxels.
 However, it is not necessarily the rule that the lowest level hierarchy exhibits the best performance, especially for local consensus methods that rely on individual supervoxel volumes to form a meaningful internal consensus.
Additionally, it is shown in Table~\ref{tab:SegTrackCompare} that solutions that are able to draw on multiple hierarchy levels (the ``All" configurations) exhibit a meaningful improvement in performance.
In general, it seems like the best segmentation performance occurs in the lower level regions, however a range of hierarchy levels should be employed to ensure the best segmentation performance.

\section{Conclusion}
\label{sec:conclude}

At the outset of our work, we proposed to answer several questions arising from the use of supervoxels for consensus-based video object segmentation.
First, we find that local consensus is imperative for the video object segmentation problem and is more useful than non-local consensus.
Second, hierarchical graph-based segmentation (GBH) supervoxel methods are the most reliable.
Finally, the lowest hierarchy levels are most effective for non-local consensus, whereas low- and mid-hierarchy levels are effective for local consensus.
In the context of video object segmentation, the highest hierarchy levels are not effective.

We find that this study is but one of many studies that have potential to shed light on the capability that supervoxels play in video understanding.
Furthermore, given the performance of the current work on accepted benchmarks, we postulate that such studies are worthwhile for the video object segmentation problem.
Given that the current work is restricted to a single object hypothesis, we are currently working on extending this analysis to a multiple object hypothesis.

Source code for the current work is provided at \cite{codeSVXGRM}.

\section*{Acknowledgements}
\noindent This work was partially supported by the DARPA MediFor program under contract FA8750-16-C-0168.

{\small
\bibliographystyle{ieee}
\bibliography{SVXConBib}
}

\end{document}